\documentclass[authoryear,preprint,5p,twocolumn,times]{elsarticle}

\usepackage{amssymb}
\usepackage{textcomp}
\usepackage{array}
\usepackage{multirow}
\usepackage{tablefootnote}
\usepackage{amsmath}
\usepackage{subcaption}
\usepackage{pbox}
\usepackage{booktabs}
\usepackage{arydshln}
\usepackage{subcaption}
\usepackage{color,soul}

\makeatletter
\def\ps@pprintTitle{%
    \let\@oddhead\@empty
    \let\@evenhead\@empty
    \def\@oddfoot{\footnotesize\itshape
         {}}%
    \let\@evenfoot\@oddfoot
    }
\makeatother

\journal{Neural Networks}

\begin{document}

\begin{frontmatter}



\title{Emotional EEG Classification using Connectivity Features and Convolutional Neural Networks}


\author[1]{Seong-Eun Moon\corref{cor2}}
\ead{se.moon@yonsei.ac.kr}
\address[1]{School of Integrated Technology, Yonsei University, Republic of Korea}

\author[3]{Chun-Jui Chen}
\ead{cjuchen@ucdavis.edu}
\address[3]{Department of Statistics, University of California, Davis, USA}

\author[4]{Cho-Jui Hsieh}
\ead{chohsieh@cs.ucla.edu}
\address[4]{Department of Computer Science, University of California, Los Angeles, USA}

\author[3]{Jane-Ling Wang}
\ead{janelwang@ucdavis.edu}

\author[1]{Jong-Seok Lee\corref{cor1}}
\ead{jong-seok.lee@yonsei.ac.kr}
\cortext[cor1]{Corresponding author}
\begin{abstract}
Convolutional neural networks (CNNs) are widely used to recognize the user's state through electroencephalography (EEG) signals. 
In the previous studies, the EEG signals are usually fed into the CNNs in the form of high-dimensional raw data. 
However, this approach makes it difficult to exploit the brain connectivity information that can be effective in describing the functional brain network and estimating the perceptual state of the user.
% There are few approaches that employed traditional EEG features and used them as input signals of the CNNs. However, the characteristic of CNNs that realizes the noticeable achievement of CNNs has not been properly considered, particularly, for the brain connectivity. 
We introduce a new classification system that utilizes brain connectivity with a CNN and validate its effectiveness via the emotional video classification by using three different types of connectivity measures. 
Furthermore, two data-driven methods to construct the connectivity matrix are proposed to maximize classification performance. 
Further analysis reveals that the level of concentration of the brain connectivity related to the emotional property of the target video is correlated with classification performance.
\end{abstract}

%\includegraphics{connMat_overall_v3_1.eps}


\end{frontmatter}


\section{Introduction}

%\IEEEPARstart{T}{he} design of a product 
The perceptual aspects of the user experience such as emotion and preference are important for various multimedia applications and services. 
For example, services such as video streaming and content recommendation can benefit from understanding the user's perception of the given multimedia content. 
It is also crucial to recognize the user's perceptual states to enhance the user's experience in virtual reality. 
In particular, emotion, which influences both our individual and social behaviors \citep{Dolan02}, is one of the most distinguishing perceptual factors. 
Therefore, many studies have tried to determine its nature \citep{Frijda88,Cabanac02,Winkielman04} and characteristics \citep{Ekman87}. 

Although emotion is traditionally investigated through explicit questionnaires or interviews, implicit measurement of emotion via physiological signals has received much attention recently 
%The interview and questionnaire directly ask users about their emotion induced by the stimulation, which brings the most reliable information. In contrast, emotion is passively observed by measuring the physiological signals in the implicit method, which gives the implicit measurement 
due to its advantages over the explicit approach.
Example modalities for physiological signal measurement include electroencephalography (EEG) \citep{Moon17survey}, functional magnetic resonance imaging (fMRI) \citep{Koelsch06}, functional near-infrared spectroscopy (fNIRS)  \citep{Bandara2018}, magnetoencephalography (MEG) \citep{Abadi15}, and peripheral physical signals \citep{Kim2008}.
The implicit measurements enable the real-time monitoring of emotion, whereas a questionnaire or interview needs to be conducted after the stimulus presentation, which leads to a lag between the emotional event and the measurement. 
Moreover, the implicit approach is relatively free from errors induced by the experimenters and the evaluation processes. 

In particular, EEG is a method to capture electrical brain activity, which is expected to contain comprehensive information on the emotional process. 
Furthermore, EEG has advantages over other cerebral physiological channels, such as high temporal resolution, low cost, and the portability of the equipment \citep{Moon17survey}. 
Thus, many studies have utilized EEG signals for emotion analysis. 

Recently, deep learning approaches have been applied to EEG signals to classify the user's emotional state. 
Most previous studies have focused on extracting representations appropriate for classification from raw EEG signals using deep learning models. 
However, EEG signals usually contain much more intense noise than the image and audio signals that the deep learning approach mostly has handled. 
Therefore, it could be beneficial to extract meaningful information first from the EEG signals and then learn it via deep learning models. 

While instant changes in amplitude or latency and spectral powers have been traditionally used to represent EEG signals, features related to brain connectivity have emerged recently because they can consider the relationship between the different brain regions \citep{Abril2018}. 
Brain connectivity has been actively employed in neuroscientific research. 
Moreover, the effectiveness of brain connectivity features in recognizing emotional states was validated in recent studies \citep{Costa06,Lee14}. 
However, brain connectivity has not been employed in deep learning approaches except in our preliminary work \citep{Moon18}. %s \citep{Moon18,Moon18evaluation} \hl{that tried to extract useful information from connectivity features for valence and liking classifications using deep learning approaches.}

In this paper, we propose a new method to exploit the connectivity information in deep learning to recognize emotional experience using EEG. 
In our method, the connectivity information is represented as a matrix obtained from the raw EEG signals. 
In particular, we utilize convolutional neural networks (CNNs) because of their capability of dealing with the spatial information of the input data. 
Furthermore, we note that as the convolution operation considers the spatial neighboring values in the input data at the same time, the performance of the proposed method varies depending on the arrangement of the connectivity matrix.
Therefore, we present two different methods to determine the arrangement of the connectivity matrix and investigate its influence on the performance of emotional video classification. 

Our contributions are summarized as follows:
\begin{itemize}
	\item We propose a CNN-based system that enables us to learn representations of neural activities based on brain connectivity by utilizing a \textit{connectivity matrix}, and we verify its effectiveness on emotional video classification.
	Several connectivity measures, namely the Pearson correlation coefficient (PCC), phase locking value (PLV), and transfer entropy (TE), are considered to construct the connectivity matrix. 
	\item We present an effective data-driven approach to arrange the connectivity matrix to maximize performance. 
	We consider two different methods for this, which utilize the similarity and dissimilarity between the EEG signals at different spatial locations, respectively. 
	\item We analyze the results of emotional video classification with respect to the emotional valence and demonstrate that the \textit{concentrativeness} of the valence-related connectivity in an input connectivity matrix is correlated with the classification performance. 
\end{itemize}

The rest of the paper is organized as follows. 
We provide a survey of the related studies in Section \ref{sec:relwork}. 
Section \ref{sec:connmat} contains a description of our proposed method and the validation of its effectiveness through experiments.
In Section \ref{sec:ord}, we discuss the influence of EEG electrode ordering on our proposed method, propose data-driven ordering methods, and provide further analysis on the concentrativeness of the valence-related connectivity.
Finally, we provide conclusions on our work in Section \ref{sec:conclusion}.

%%%%%%%%%%%%%%%%%%%%%%%%%%%%%%%%%%%

\section{Related Work}
\label{sec:relwork}

In this section, we review existing studies to analyze emotion using EEG signals, which are categorized depending on the type of EEG features and learning models. 
Representative studies are summarized in Table \ref{tab:relwork}.

\begin{table*}[ht!]
	\centering
	\caption{Results of emotion classification reported in representative existing studies.}~\label{tab:relwork}
	\small
	\begin{tabular}{>{\centering}m{2.5cm} >{\centering}m{4cm} >{\centering}m{3cm} >{\centering}m{2cm} >{\centering}m{4.5cm}}
		\toprule \midrule
		& Classification target & Input data & Classifier & Classification accuracy \tabularnewline \midrule		
		\cite{Liu18study} & disgusting vs. happy vs. neutral vs. sad vs. tense & PSD and asymmetry features & SVM & within-stimulus: 93.3\% (PSD), 85.4\% (asymmetry), \\ cross-stimulus: 68.9\% (PSD), 64.4\% (asymmetry) \tabularnewline \midrule
		\cite{Liu18realtime} & positive emotions (joy vs. amusement vs. tenderness), negative emotions (anger vs. disgust vs. fear vs. sadness) & PSD and asymmetry features & SVM & positive emotions: 86.4\%, \\ negative emotions: 65.1\% \tabularnewline \midrule
		\cite{Mert18} & high- vs. low-valence,\\ high- vs. low-arousal & time-frequency features obtained by the multivariate synchrosqueezing transform & fully connected neural network & valence: 82.0\%, \\ arousal: 82.1\% \tabularnewline 
		\midrule
		\cite{Zhang16} & low-valence \& low-arousal vs. low-valence \& high-arousal vs. high-valence \& low-arousal vs. high-valence \& high-arousal & time-frequency features using empirical mode decomposition & SVM & 93.2\% \tabularnewline \midrule
		\cite{Yanagimoto16} & positive vs. negative & raw EEG signals & CNN & 73.9\% \tabularnewline \midrule
		\cite{Alhagry17} & low- vs. high-\\valence/arousal/liking & raw EEG signals & LSTM & valence: 85.4\%, \\ arousal: 85.6\%, \\ liking: 88.0\% \tabularnewline \midrule
		\cite{Bozhkov2016} & positive vs. negative & raw EEG signals & echo state network +SVM & 98.1\% \tabularnewline \midrule 
		\cite{Zheng15} & positive vs. neutral vs. negative & differential entropy features & deep belief network & 86.6\% \tabularnewline \midrule
		\cite{Li18} & positive vs. neutral vs. negative & differential entropy features & CNN & 83.8\% \tabularnewline \midrule
		\cite{Li17} & low-valence \& low-arousal vs. low-valence \& high-arousal vs. high-valence \& low-arousal vs. high-valence \& high-arousal & PSD & hybrid model of LSTM and CNN & 75.2\% \tabularnewline \midrule
		\cite{Lee14} & positive vs. neutral vs. negative & correlation, coherence, and phase synchronization features & quadratic discriminant analysis & correlation: 61\%, \\ coherence: 62\%, \\ phase synchronization: 82\% \tabularnewline \midrule
		\cite{Clerico15} & low- vs. high-\\valence/arousal/liking/dominance & mutual information+PSD+asymmetry features & SVM & valence: 58\%, \\ arousal: 66\%, \\ liking: 64\%, \\ liking: 62\% \tabularnewline \midrule
		%		\citep{Wyczesany18} & happy vs. angry & DTF features & & \tabularnewline \midrule
		\cite{Shahabi16} & joyful vs. neutral,\\joyful vs. melancholic & DTF features & SVM & joyful vs. neutral: 93.7\%, \\ joyful vs. melancholic: 80.4\% \tabularnewline
		
		\bottomrule
	\end{tabular}
	\raggedright\\
	\vspace{0.2cm}
	{\scriptsize
	PSD: power spectral density, SVM: support vector machine, CNN: convolutional neural network, LSTM: long short-term memory, DTF: directed transfer function
	}
\end{table*}

\subsection{EEG-based emotion analysis}
\label{sec:relwork-emotion}

The response of the brain to emotional events has been actively explored through EEG. 
Typically, the EEG signals are represented by features, which can be categorized depending on the domain that they consider \citep{Jenke14}. 

Time domain features include statistical features such as the mean, standard deviation, power, Hjorth features \citep{Hjorth70}, length density, fractal dimensions, etc. % 
In particular, the instant phase and amplitude changes of EEG signals (called event-related potential) have been investigated to discover the brain responses to emotional stimuli in many studies \citep{Olofsson08}. 
They showed that the valence of emotion, which ranges from negative (low-valence) to positive (high-valence) levels, tends to manipulate the early EEG signals, % (100-250 ms),
and the influence of arousal describing the intensity of emotion appears relatively later. 

The power of EEG signals in the frequency domain is one of the most popular EEG features for emotion analysis.
Typically, high-amplitude signals in the low-frequency range are observed when subjects are in a calm state and, in contrast, high-amplitude signals in the high-frequency range are evident in an alert state \citep{Bear15}. 
Power features are also used to derive other frequency domain features.
For example, the asymmetry index, which is calculated as the difference between the power values obtained from symmetrically located electrode pairs, is utilized to describe the asymmetry between the left and right hemispheres of the brain.
In \cite{Liu18study}, the frequency domain features were used to classify discrete emotional states (disgust, happiness, neutrality, sadness, and tenseness) induced by watching a video, where the asymmetry indexes of the 12 electrode pairs were employed in addition to the power spectral density (PSD) of the EEG signals. 
\cite{Liu18realtime} used the same features to distinguish positive emotions (joy, amusement, and tenderness) and negative emotions (anger, disgust, fear, and sadness) induced by watching videos. 

For the frequency domain features, it is assumed that the EEG signals are stationary for the duration of a trial, but such an assumption may not hold in some cases. 
Therefore, the features that consider the information in the time and frequency domains jointly have been proposed.  
For example, \cite{Mert18} performed classification between the binary states (high vs. low) of arousal and valence using the time-frequency distribution of EEG signals obtained by the multivariate synchrosqueezing transform. 
Meanwhile, \cite{Zhang16} applied empirical mode decomposition, which is a time-frequency analysis method based on the Hilbert-Huan transformation, for emotion classification in the valence-arousal plane.

The aforementioned features are mostly used with conventional shallow machine learning models. 
Deep learning-based emotion analysis methods capable of extracting more effective features through learning have been recently proposed in the literature.

\subsection{Deep learning approaches for emotion recognition}
\label{sec:relwork-deep}

Deep learning approaches to recognize emotional states can be roughly categorized into two types: 
(1) extracting features from raw EEG signals and (2) utilizing refined information from EEG signals instead of raw signals to derive further representative features.

The former is a completely data-driven method and thus has the potential to maximize performance with a proper learning model and scheme. 
For example, \cite{Yanagimoto16} employed raw EEG signals for binary classification of emotional valence by using CNNs, which outperformed shallow models such as support vector machine (SVM) and random forest. 
Furthermore, recurrent neural networks (RNNs) were applied to raw EEG signals for emotional state classification by \cite{Alhagry17} (high- vs. low-arousal, positive vs. negative valence, and high- vs. low-liking) and \cite{Bozhkov2016} (high- vs. low-valence). 

However, EEG signals are highly complex because they reflect various aspects of perceptual experience in addition to emotion. 
Moreover, there are differences in EEG patterns among different individuals, which imposes further difficulty in extracting meaningful emotion-related representation from raw EEG signals via learning.
Therefore, it can be helpful to first process the raw EEG signals and then utilize the results for further feature extraction using deep learning models, which is the basis of the second approach and inspired our study.

\cite{Zheng15} used differential entropy, which is a measure of the amount of information included in EEG signals, as the input of deep belief networks for the classification of emotional states (positive, neutral, and negative). 
Differential entropy was also employed by \cite{Li18} to distinguish the positive, neutral, and negative emotional states induced by videos, where the CNN receives input as a topographical 2D image representation of the differential entropy based on the spatial arrangement of the EEG electrodes.
Meanwhile, \cite{Li17} designed a hybrid model incorporating RNN and CNN for emotion classification in the valence-arousal plane by using topographies of the power spectral densities (PSDs) of the EEG signals. 

The EEG features employed in CNN-based studies (i.e., differential entropy and PSD) represent the activity in each brain region. 
However, it is generally known that the brain regions consist of a network, and that brain functions can be interpreted as interactions between the regions through the network \citep{Hassan18}. 
Therefore, emotion analysis can also benefit from examining the relationship between different brain regions, which motivated us to employ brain connectivity from EEG signals.

\subsection{Connectivity-based EEG analysis}
\label{sec:relwork-conn}

\begin{figure*}[h]
	\centering
	\includegraphics[width=\textwidth]{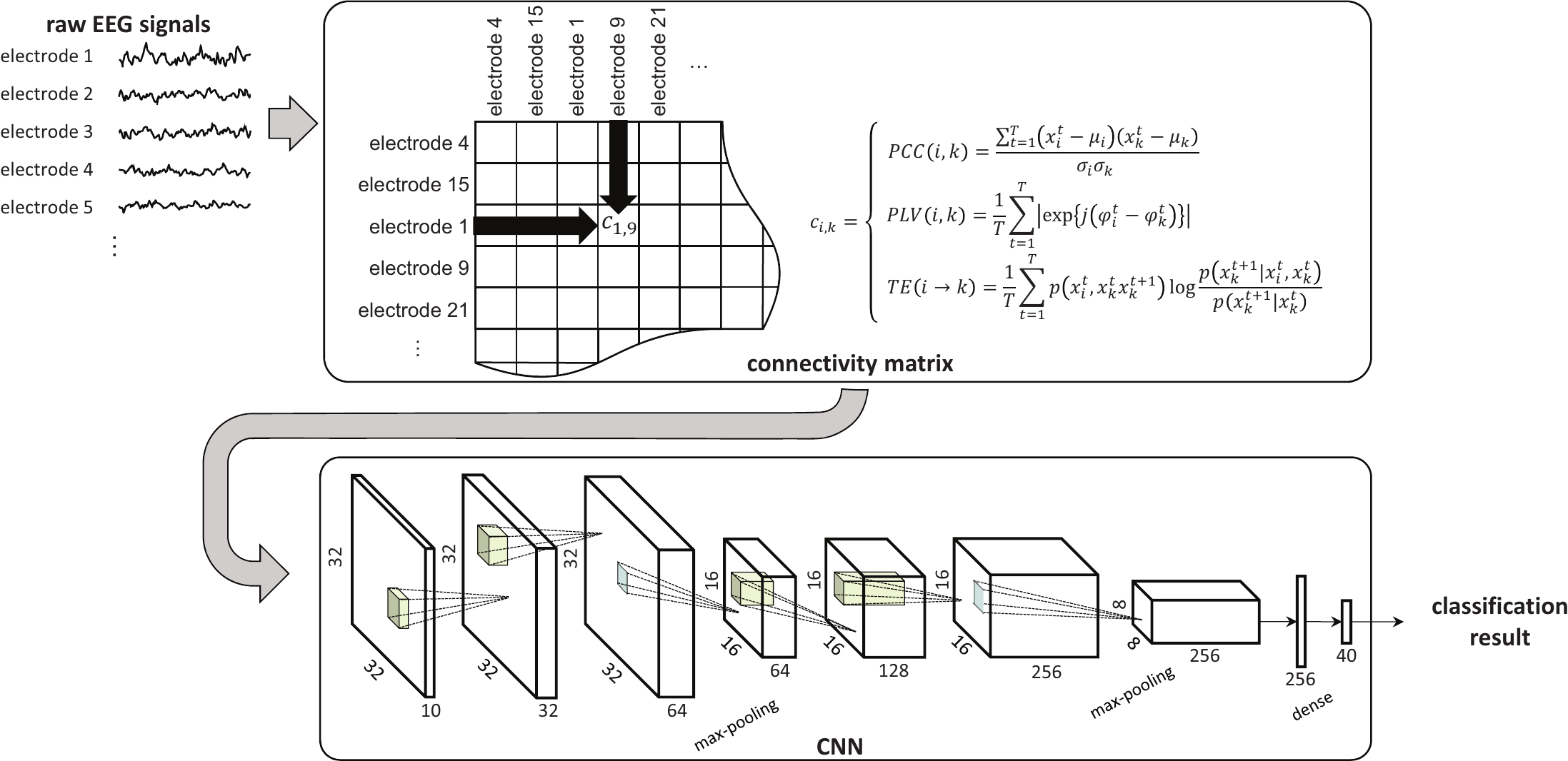}
	\caption{Overview of the proposed classification system.}
	\label{fig:connMat}
\end{figure*}

While conventional EEG features consider different brain regions individually, brain connectivity examines the relationship between brain regions by measuring the dependencies of brain activity such as coactivation and causal relationships. 
Therefore, brain connectivity provides information about brain activity from a different perspective than with conventional EEG features. 

Several studies have demonstrated the effectiveness of brain connectivity for emotion recognition. 
\cite{Costa06} revealed that phase synchronization between EEG signals is influenced by positive and negative emotions induced by watching videos. 
\cite{Lee14} employed three types of connectivity measures (correlation, coherence, and phase synchronization) to distinguish positive, neutral, and negative emotional states induced by watching emotional videos. 
The mutual information of inter-hemispheric pairs was used for binary classification of emotional states (high vs. low for valence, arousal, liking, and dominance) by \cite{Clerico15}. 
\cite{Wyczesany18} used the directed transfer function (DTF), which is a measure of the causal relationship based on Granger causality, to investigate emotional responses to happy and angry face images. 
The DTF was also employed to analyze emotional states (joyful, neutral, and melancholic) by \cite{Shahabi16}. 

Existing emotion analyses using the brain connectivity of EEG signals have been limited to statistical tests \citep{Costa06,Lee14,Wyczesany18} or conventional machine learning methods \citep{Lee14,Clerico15,Li19,Shahabi16}.
To the best of our knowledge, there have not been any attempts to use brain connectivity with deep learning, which is probably due to the absence of a proper way to utilize brain connectivity features within deep learning models. 
In this study, we introduce the connectivity matrix, which is a method to represent the connectivity information for learning with CNNs, and evaluate its effectiveness.

%%%%%%%%%%%%%%%%%%%%%%%%%

\section{Proposed System}
\label{sec:connmat}

Figure \ref{fig:connMat} presents a summary of the proposed classification system to utilize brain connectivity with a CNN. 
After the connectivity measures are calculated from the raw EEG signals, the connectivity matrix is constructed based on a certain ordering method. 
The CNN receives the connectivity matrix as an input signal and is trained to extract meaningful representations for the target classification task.

\subsection{EEG connectivity matrix}
\label{sec:connmat-conn}

Three different connectivity measures are used in this study: the Pearson correlation coefficient (PCC), phase locking value (PLV), and transfer entropy (TE), which have been popularly employed in neuroscientific studies and reflect various aspects of brain connectivity. 

The PCC measures the linear relationship between two signals as a continuous number ranging from -1 to 1. 
PCC values of -1 and 1 correspond to perfect negative and positive linear relationships, respectively, and a PCC value of zero indicates that the two signals are uncorrelated. 
Let $\mathbf{x}_i=\{x_i^1, x_i^2, ..., x_i^T\}$ denote an EEG signal of the $i$-th electrode, where $T$ is the time length of the signal. 
The PCC of two signals $\mathbf{x}_i$ and $\mathbf{x}_k$ is calculated as

\begin{equation}
    PCC(i,k) = \frac{\frac{1}{T}\sum_{t=1}^{T}(x_i^t-\mu_i)(x_k^t-\mu_k)}{\sigma_i\sigma_k},
    \label{eq:pcc}
\end{equation}
where $\mu$ and $\sigma$ are the mean and standard deviation of the signal, respectively.

The PLV \citep{Lachaux99} describes the phase synchronization between two signals, which is calculated by averaging the absolute phase differences as follows:

\begin{equation}
    PLV(i,k) = \frac{1}{T}\left\vert \sum_{t=1}^{T} \exp{ \left\{ j(\varphi_i^t-\varphi_k^t) \right\} }\right\vert,
    \label{eq:plv}
\end{equation}
where $\varphi^t$ is the phase of the signal at time $t$. 
It ranges from 0 to 1, indicating that the two signals are either perfectly independent or perfectly synchronized, respectively.

The TE \citep{Schreiber00} measures the directed flow of information from a signal $\mathbf{x}_i$ to another signal $\mathbf{x}_k$:

\begin{equation}
TE(i\to k) = \frac{1}{T-1}\sum_{t=1}^{T-1}{p(x_i^t,x_k^t,x_k^{t+1})}\log \frac{p(x_k^{t+1}|x_i^t,x_k^t)}{p(x_k^{t+1}|x_k^t)}.
\label{eq:te}
\end{equation}
In other words, it describes the gain obtained by knowing $\mathbf{x}_i$ for the prediction of $\mathbf{x}_k$.
A TE value of zero means that there is no causal relationship between the two time series.
The Java Information Dynamics Toolkit \citep{Lizier14} is used to calculate TE features in this study.

The connectivity features are calculated for every pair of EEG electrodes. Therefore, if there are $N_e$ electrodes, the number of obtained features is $N_e(N_e-1)/2$ for undirected connectivity (PCC or PLV) or $N_e(N_e-1)$ for directed connectivity (TE).

The connectivity features for all electrode pairs can be represented as a matrix in which the element at $(i,k)$ indicates the connectivity between the EEG signals obtained from the $i$-th and $k$-th electrodes, as shown in Figure \ref{fig:connMat}. 
This \textit{connectivity matrix} is equivalent to the adjacency matrix of a graph in which the EEG electrodes are considered as nodes and the connectivity features as edge weights.

One issue arising here is node ambiguity, i.e., how to order the electrodes in the connectivity matrix. 
The geometry of the electrodes is used for the ordering in this section, and data-driven methods are described in Section \ref{sec:ord}.

\begin{figure}[t]
    \centering
    \begin{subfigure}{0.3\textwidth}
        \includegraphics[width=\columnwidth]{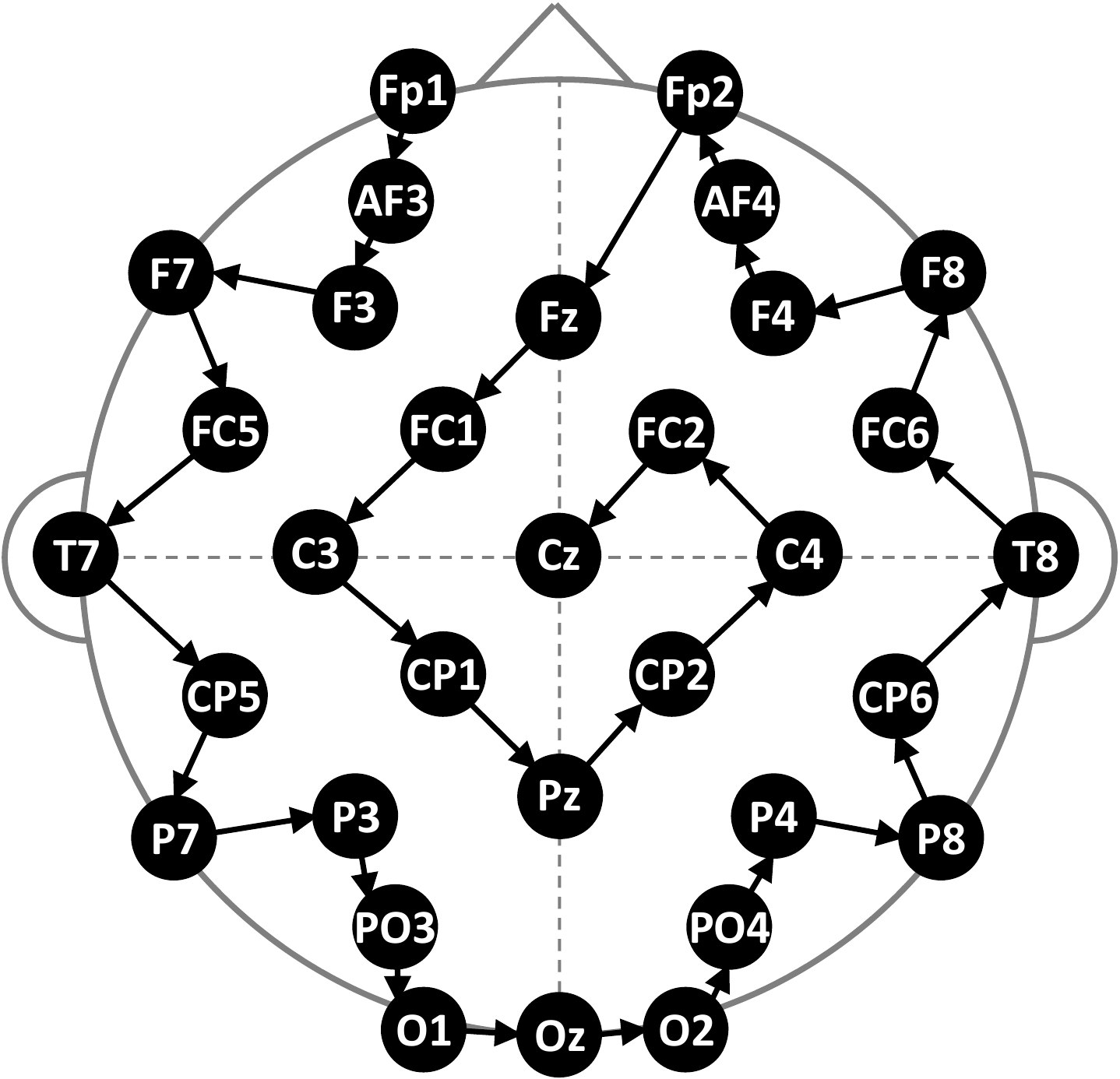}
        \caption{\textsf{dist}}
        \label{fig:order-closest1}
    \end{subfigure}
    
    \vspace{0.5cm}
    
    \begin{subfigure}{0.3\textwidth}
        \includegraphics[width=\columnwidth]{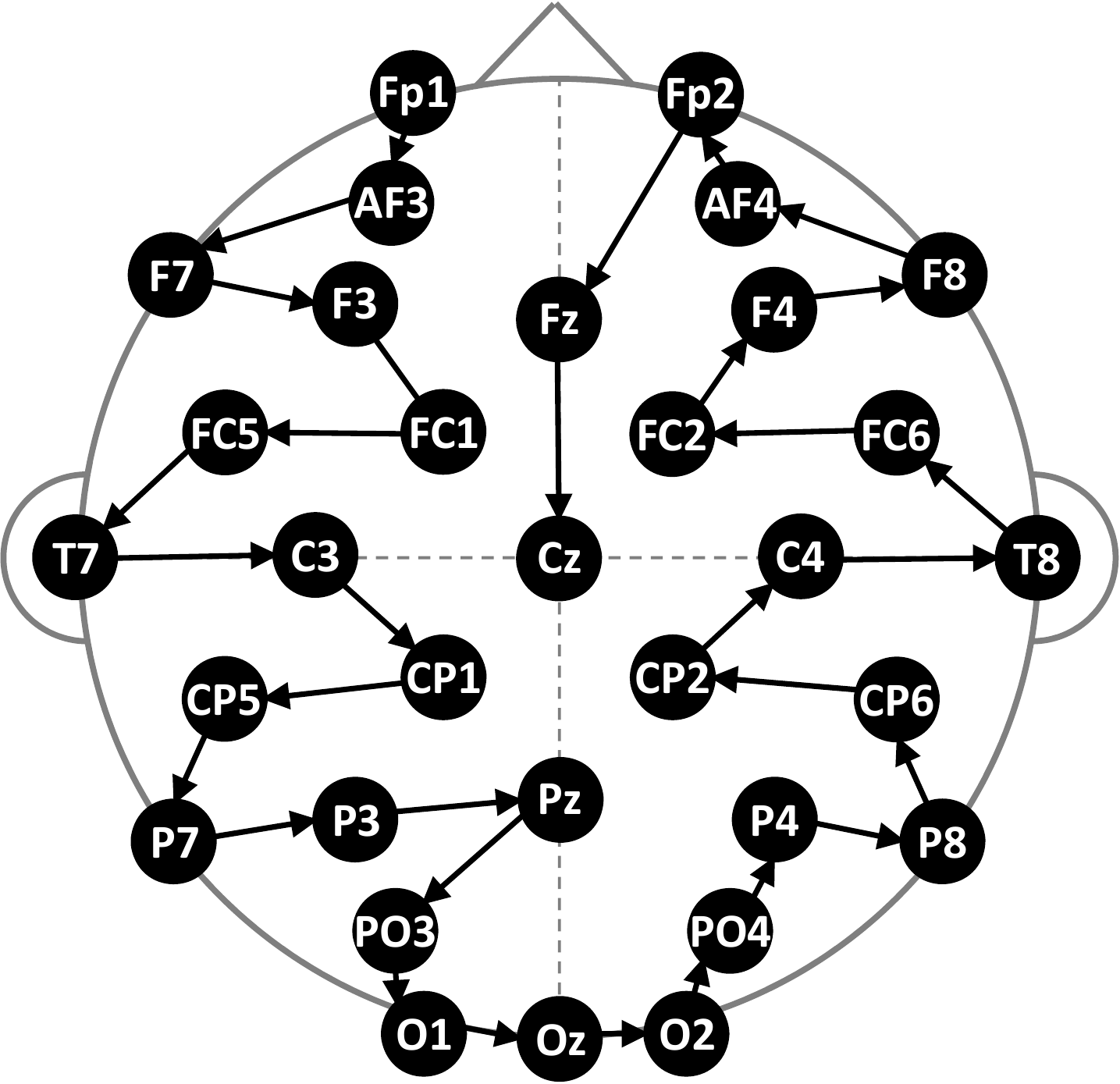}
        \caption{\textsf{dist-restr}}
        \label{fig:order-closest2}
    \end{subfigure}
    \caption{Distance-based ordering methods used to construct connectivity matrices.}
    \label{fig:dist-order}
\end{figure}

Determining the order based on the locations of the electrodes on the scalp is to consider that the EEG signals obtained from neighboring brain regions tend to be similar due to the volume conductance effect \citep{Broek98}, which enables the construction of a smooth connectivity matrix. 
Specifically, starting from the electrode on the left-frontal region, the one that is the closest to the current electrode is selected as the next electrode. 
The result of this ordering for a 32-channel EEG system is shown in Figure \ref{fig:order-closest1}.

On the other hand, it is generally accepted that the asymmetry between the left and right hemispheres of the brain is closely related to emotional valence processes \citep{Coan04,Reznik18}. 
Therefore, we introduce another ordering method to describe the hemispheric asymmetry. 
This also starts from the left-frontal region and proceeds to the closest electrode, but the candidates for the next electrode are limited to the electrodes in the same hemisphere. 
The ordering trajectory can cross the hemispheric border only when there are no available electrodes in the same hemisphere. 
Figure \ref{fig:order-closest2} illustrates the result of this ordering method. 
These two ordering methods are denoted as \textsf{dist} and \textsf{dist-restr}, respectively.

\subsection{Database and classification problem}
\label{sec:connmat-data}

\begin{figure}[t]
    \centering
    \includegraphics[width=0.35\textwidth]{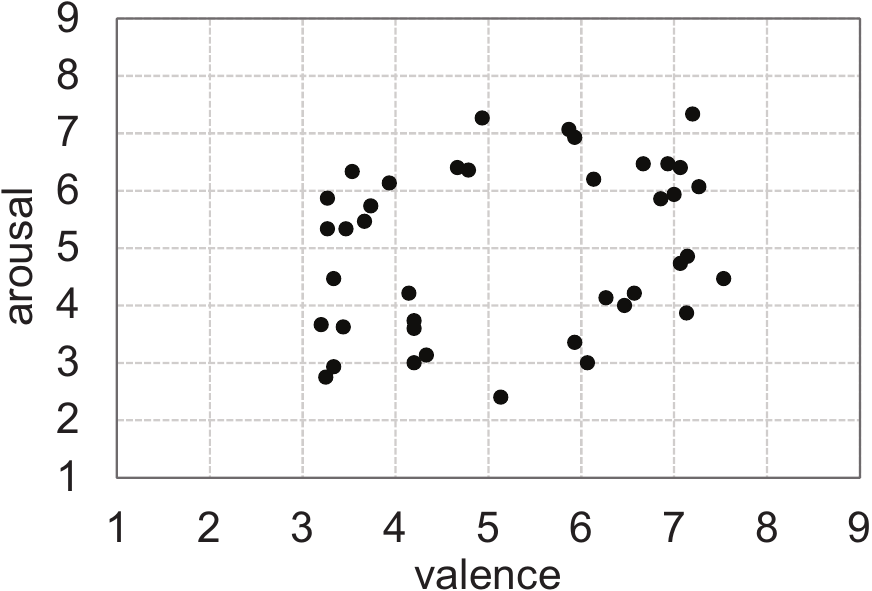}
    \caption{Emotional characteristics of the videos in DEAP \citep{Koelstra09} in terms of valence and arousal scores (averaged across the subjects).}
    \label{fig:va-video}
\end{figure}

DEAP (Database for Emotion Analysis using Physiological Signals) by \cite{Koelstra12}, which is one of the largest EEG databases for emotion analysis, is employed in this study.
It contains 32-channel EEG signals from 32 subjects captured while the subjects were watching 40 emotional music videos. 
In addition, subjective scores that quantify the levels of valence, arousal, liking, and dominance of the emotional states induced by watching the videos are included in the database. 
Figure \ref{fig:va-video} shows the videos in the valence-arousal plane. 
We use the preprocessed EEG signals provided in the database that had undergone downsampling to 128 Hz, removal of eye movement artifacts using a blind source separation technique, and band-pass filtering from 4 to 45Hz.

We apply band-pass filtering to the EEG signals to extract the delta (0-3 Hz), theta (4-7Hz), low-alpha (8-9.5 Hz), high-alpha (10.5-12 Hz), alpha (8-12 Hz), low-beta (13-16 Hz), mid-beta (17-20 Hz), high-beta (21-29 Hz), beta (13-29 Hz), and gamma (30-50 Hz) sub-frequency bands. 
The connectivity matrices obtained for these 10 sub-frequency bands are stacked along the depth axis, and so the size of the input data becomes 32 $\times$ 32 $\times$ 10 (number of electrodes $\times$ number of electrodes $\times$ number of sub-frequency bands).
We divide the EEG signals into three-second-long segments with an overlap of 2.5 seconds to obtain a sufficient number of data samples for training the CNN. 
Since a single trial of the database is one minute long, we obtain 115 EEG signal segments for each trial. 

The emotional video classification task as defined by \cite{Jang18} is considered for the experiment.
The EEG data are randomly divided into training, validation, and test data, which hold 80\%, 10\%, and 10\% of the entire data, respectively, as in \cite{Jang18}. 
The validation data are used to determine when the learning process needs to be stopped and to select the best CNN.

\subsection{Classifier}
\label{sec:connmat-cls}

\begin{table}[t]
	\centering
	\caption{CNN architecture.}~\label{tab:cnn}
	\small
	\begin{tabular}{r c c c c}
	    \toprule \midrule
	    Layer & Type         & Output shape            & Kernel size & Stride \\ \midrule
	    1     & convolution  & 32$\times$32$\times$32  & 3 or 5        & 1 \\
	    2     & convolution  & 32$\times$32$\times$64  & 3 or 5        & 1 \\
	    3     & max-pooling  & 16$\times$16$\times$64  & 2$\times$2  & 2 \\
	    4     & convolution  & 16$\times$16$\times$128 & 3 or 5        & 1 \\
	    5     & convolution  & 16$\times$16$\times$256 & 3 or 5        & 1 \\
	    6     & max-pooling  & 8$\times$8$\times$256   & 2$\times$2  & 2 \\
	    7	  & dense        & 256                     & -           & - \\
	    8     & softmax      & 40                      & -           & - \\ 
	    \bottomrule
	\end{tabular}
\end{table}

Our CNN consists of four convolutional layers with rectified linear unit (ReLU) activation functions, two max-pooling layers, and a fully connected layer, as detailed in Table \ref{tab:cnn}. 
We employ two different convolution kernel sizes ($s$=3 and 5) and evaluate the influence of the kernel size on the classification performance. 
The dropout with a probability of 0.5 is applied to the fully connected layer and the output layer, and batch normalization is used after each max-pooling layer. 

The Adam algorithm \citep{Kingma2014} is used for training, and the cross-entropy is employed as the loss function of learning. 
The learning rate is initially set to 0.0001 and decreased by 0.8 times at every 10 epochs. 
The training is stopped when the validation loss does not decrease for 40 epochs, and the test is conducted using the network that shows the best validation accuracy. 
The CNN model and classification process are implemented in PyTorch. 

\subsection{Results}
\label{sec:connmat-result}

\begin{table}[t]
	\centering
	\small
	\caption{Results of the emotional video classification in terms of accuracy. $s$ indicates the convolution kernel size of CNNs. }~\label{tab:result1}
	\begin{tabular}{c l l c}
		\toprule \midrule
		                          & Method &                                          & Accuracy (\%)\\ \midrule
		\multirow{14}{*}{Baseline} & \multicolumn{2}{l}{k-nearest neighbors \citep{Jang18}}& 48.50 \\
		                          & \multicolumn{2}{l}{k-nearest neighbors (PCC)} & 33.48 \\
		                          & \multicolumn{2}{l}{k-nearest neighbors (PLV)} & 44.55 \\
		                          & \multicolumn{2}{l}{k-nearest neighbors (TE)}  & 38.25 \\ 
		                          
		                          & \multicolumn{2}{l}{random forest \citep{Jang18}}      & 51.34 \\
		                          & \multicolumn{2}{l}{random forest (PCC)} & 49.60 \\
		                          & \multicolumn{2}{l}{random forest (PLV)} & 42.75 \\
		                          & \multicolumn{2}{l}{random forest (TE)}  & 20.26 \\
		                          
		                          & \multicolumn{2}{l}{neural network (PCC)} & 16.95 \\
		                          & \multicolumn{2}{l}{neural network (PLV)} & 8.77 \\
		                          & \multicolumn{2}{l}{neural network (TE)}  & 10.75 \\
		                          
		                          & \multicolumn{2}{l}{PSD+CNN ($s$=3) \citep{Li17}} & 31.01 \\
		                          & \multicolumn{2}{l}{PSD+CNN ($s$=5) \citep{Li17}} & 37.18 \\ 
		                          & \multicolumn{2}{l}{graph CNN \citep{Jang18}} & 65.27 \\ \midrule
		\multirow{12}{*}{Proposed} & \multirow{4}{*}{PCC} & \textsf{dist} ($s$=3) & 66.96 \\
		                           &                      & \textsf{dist} ($s$=5) & 71.96 \\
		                           &                      & \textsf{dist-restr} ($s$=3) & 67.67 \\
		                           &                      & \textsf{dist-restr} ($s$=5) & 74.53 \\ \cmidrule{2-4}
		                           & \multirow{4}{*}{PLV} & \textsf{dist} ($s$=3) & 72.09 \\
		                           &                      & \textsf{dist} ($s$=5) & 80.73 \\
		                           &                      & \textsf{dist-restr} ($s$=3) & 73.12 \\
		                           &                      & \textsf{dist-restr} ($s$=5) & 75.01 \\ \cmidrule{2-4}
		                           & \multirow{4}{*}{TE}  & \textsf{dist} ($s$=3) & 58.64 \\
		                           &                      & \textsf{dist} ($s$=5) & 65.06 \\
		                           &                      & \textsf{dist-restr} ($s$=3) & 55.42 \\
		                           &                      & \textsf{dist-restr} ($s$=5) & 65.44 \\
		\bottomrule
	\end{tabular}
\end{table}

%We compare our method with several baseline methods. 
%As traditional classifiers, the k-nearest neighbors method using entropy features and the random forest method using signal power features are used \citep{Jang18}. 
%We also test the k-nearest neighbors and random forest methods using connectivity features (i.e., PCC, PLV, and TE).
%In addition, a neural network with two fully connected hidden layers having 512 and 256 nodes using the connectivity features is examined.
%As a simple deep learning approach, a CNN takes 2D PSD topographies \citep{Li17} as input, which does not consider the connectivity. 
%In the graph CNN approach \citep{Jang18}, a graph is constructed based on the distances between electrodes and the signal entropy is used as the features on the graph, which are modeled by the graph CNN \citep{Defferrard16}; this approach implicitly considers the connectivity information underlying in the given graph, which is fixed for all data.

We compare our method with several traditional methods: the k-nearest neighbors method using entropy features \citep{Jang18} or  connectivity features (PCC, PLV, or TE), the random forest method using signal power features \citep{Jang18} or the connectivity features, and a neural network with two fully connected hidden layers having 512 and 256 nodes using the connectivity features.
In addition, we compare our method with a CNN receiving 2D PSD topographies as input \citep{Li18}, which does not consider the connectivity, and the graph CNN approach by \cite{Jang18} using the graph constructed based on the distances between the electrodes and the signal entropy features on the graph, which implicitly considers the underlying connectivity information in the given graph but assumes a fixed connectivity structure for all data.

The classification results are reported in Table \ref{tab:result1}.
The obtained classification accuracies of the proposed method are significantly higher than random chance (i.e., 2.5\%), the accuracies of the traditional classifiers (k-nearest neighbors, random forest, and neural network), and the accuracies of the CNN approach without considering the connectivity (PSD+CNN). 
One-sample Wilcoxon signed-rank tests under the null hypothesis that the median of the accuracies of the proposed method is the same to the accuracy of a baseline method confirm the significance of the superiority of the proposed method ($p<0.05$ for all comparisons). 
In particular, the traditional classifiers using the connectivity features are not so effective, which shows that the classifiers do not efficiently exploit the connectivity features. 
Among them, the neural networks show the worst performance. As the connectivity features are high-dimensional, the number of parameters included in the neural networks is large compared to the number of training data, which probably results in overfitting of the neural networks.
Moreover, our method (with PCC or PLV) is superior to the graph CNN approach that also considers the connectivity information (but not in a data-specific way), which is confirmed by a one-sample Wilcoxon signed-rank test ($p<0.01$).
This demonstrates that explicitly exploiting the connectivity information specific to each sample is beneficial.

The classification performance is enhanced by increasing the kernel size. 
The performance improvement with $s$=5 compared with $s$=3 is consistently observed for all connectivity features, indicating that aggregating the input signal information over a relatively wide area through convolution is beneficial. 
However, we observed that a larger kernel size ($s$=7) did not further enhance the accuracy, which is probably because of the burden of the increased number of parameters to be trained and the over-smoothing effect of the larger kernel size.

Among the connectivity measures, PLV shows the best performance for all combinations of the ordering methods and kernel sizes, which implies the importance of phase information of EEG signals for emotional analysis. 
In contrast, the lowest accuracies are obtained using TE in all cases, one of the possible reasons being the parameter selection of TE. 
We considered the first-order TE without a time delay (i.e., only the signals at $t$ and $t+1$ are used in (\ref{eq:te})) to make it comparable to the other connectivity measures. 
Therefore, although TE attains comparable or better results compared with the baselines, additional performance improvement is expected by giving more freedom to select the parameters.

There is no significant difference between the classification performances of the two ordering methods. 
The \textsf{dist-restr} ordering method shows slightly better performances for PCC but not for PLV and TE. 
We think that, while the hemispheric asymmetry is useful for distinguishing the positiveness and negativeness of emotional states, much more diverse emotional aspects are involved in emotional video classification.

%%%%%%%%%%%%%%%%%%%%%%%%%%%%%
\section{Data-Driven Ordering of Connectivity Matrix}
\label{sec:ord}

\subsection{Influence of ordering}
\label{sec:ord-influ}

In general, the ordering of the connectivity matrix influences the results of convolution operations because the neighboring connectivity values are considered at the same time within a convolutional filter. 
For example, if the size of the convolution kernels is 3$\times$3, the output of the convolution operation at $(i,k)$ is calculated by using the element at $(i,k)$ along with its 8-connected elements.

Furthermore, the connectivity matrix must be robust to the task-irrelevant variation of EEG signals, which can appear even for the same emotional state depending on the context of stimulation, prior experience, the content of the stimuli, and so on. 
This can be achieved if the neighboring elements in the connectivity matrix have similar functional meanings so that the variation in brain responses appears at most as local translations of patterns that can be effectively managed by the CNN.

Therefore, the arrangement of the connectivity matrix needs to be carefully considered, particularly because there is no inherent axis for determining the ordering. 
This is a new challenge that is not present in other fields. 
For instance, the spatial arrangement of images is determined based on the physical structure of the objects that the pixels describe, and the arrangement of the spectrograms of audio signals is defined by the time and frequency. 
While we used the physical locations of the EEG electrodes in the arrangement in Section \ref{sec:connmat-conn}, the physical distance between the EEG electrodes is an approximated measure of the signal similarity. 
In other words, the distance-based ordering is somewhat reasonable but not optimal when determining the adjacency between the electrodes.

Two aspects of EEG signals can be considered for the ordering of the connectivity matrix: (1) collaboration across multiple regions and (2) interactions between a specific pair of regions. 
The former can be considered as a global pattern of brain connectivity and the latter as a local pattern. 
In the following, we propose data-driven approaches to describe such connectivity patterns.

\subsection{Data-driven ordering methods}
\label{sec:ord-str}

Inspired by \cite{ChenCM11} and \cite{Chen19}, we use unidimensional scaling (UDS) to determine the ordering that conveys either global or local features of the connectivity. 
UDS projects given multidimensional data points onto a unidimensional space while preserving the relative distance between them as much as possible. The distance between data points is defined by a so-called disparity function depending on the desired property of the ordering.

In order to obtain the order that enhances global features (noted as \textsf{data-global}), the electrodes showing high connectivity are placed close together. 
For this, the disparity function used in UDS should be opposite to the connectivity measure. 
Therefore, we can define the disparity function for \textsf{data-global} as
\begin{equation}\label{eq:smooth}
\delta(i,k) = 2\left(1-c_{i,k}\right),
\end{equation}
where $c_{i,k}$ represents the connectivity measure between the $i$-th and $k$-th electrodes. 
For example, when the PLV is used as the connectivity measure, the disparity function results in 0 for signals in which the phases are perfectly synchronized (i.e., the PLV value is 1), and a disparity value of 2 is obtained for perfectly independent signals. 
This disparity function is plugged into the objective function of UDS (known as the normalized stress function \citep{De1977}), which can be written as
\begin{equation}\label{eq:uds}
\mathrm{stress}(l_1,\cdots,l_{N_e}) = \frac{\sum_{i<k} \left(|l_i - l_k| - \delta(i,k) \right)^2}{\sum_{i<k} \delta(i,k)^2},
\end{equation}  
where $|l_i - l_k|$ indicates the Euclidean distance between the $i$-th and $k$-th electrodes in the projected unidimensional space.

The continuous-valued solution $(l_1,\cdots,l_{N_e})$ is obtained by minimizing the objective function so that the disparity function value and the distance in the unidimensional space become as similar as possible on average. 
We then discard the distance information and keep only the order from the solution. 
The EEG electrodes are arranged in each of the horizontal and vertical directions of the connectivity matrix according to the new order. 
This ordering process is implemented by using the regular multidimensional scaling package in MATLAB.

We can define another disparity function to obtain the order for local features as follows:
\begin{equation}\label{eq:rough}
\delta(i,k) = c_{i,k}^2.
\end{equation}
In the case of PLV, this produces disparity values of 0 and 1 for independent and perfectly synchronized signal pairs, respectively. 
Once again, the normalized stress function (\ref{eq:uds}) using this disparity function is minimized and the order information of the solution is obtained. 
In the connectivity matrix arranged through this ordering, the brain regions having strong positive or negative connectivity are separated as far as possible, and regions having zero association are arranged as closely as possible. 
This connectivity matrix enhances the local patterns of connectivity, and so this ordering method is denoted as \textsf{data-local}.

\subsection{Results}
\label{sec:ord-result}

\begin{table}[t]
	\centering
	\small
	\caption{Accuracies (\%) of the emotional video classification by using the data-driven ordering methods in comparison with the results of the distance-based ordering methods.}~\label{tab:result2}
	\begin{tabular}{r c c c c}
		\toprule \midrule
		\multirow{2}{*}{Ordering} & \multicolumn{2}{c}{PCC} & \multicolumn{2}{c}{PLV} \\ 
		            & $s$=3   & $s$=5   & $s$=3   & $s$=5 \\ \cmidrule{2-5}
		\textsf{dist}        & 66.96 & 71.96 & 72.09 & 80.73 \\
		\textsf{dist-restr}  & 67.67 & 74.53 & 73.12 & 75.01 \\ 
		\textsf{data-global} & 71.51 & 80.28 & 75.88 & 87.36 \\
		\textsf{data-local}  & 69.11 & 78.97 & 74.21 & 84.26 \\ 
		\bottomrule
	\end{tabular}
\end{table}

Table \ref{tab:result2} reports the classification results of the proposed ordering methods for PCC and PLV (TE is excluded here because it showed significantly worse classification performance than the others in Section \ref{sec:connmat}). 
The best classification accuracy of 87.36\% is obtained using \textsf{data-global} in the case of PLV and a kernel size of 5. 
The superiority of PLV and the larger kernel size is consistently observed for the connectivity matrices arranged by the data-driven ordering, as in the results of the distance-based ordering. 
The connectivity matrices arranged by the data-driven ordering methods yield significantly better classification performance compared to those based on the distance between the electrodes. 
The best accuracy is improved by approximately 7\% by adopting the data-driven ordering methods. 
This demonstrates that the connectivity matrices arranged based on the data are more appropriate to extract useful features for the emotional video classification using CNNs.
Between the two data-driven ordering methods, \textsf{data-global} shows higher classification accuracies than \textsf{data-local} in all cases, implying that the global patterns of connectivity described by \textsf{data-global} are more effective for emotional video classification.

\begin{table}[t!]
	\centering
	\small
	\caption{$\chi^2$-values of the McNemar tests for PCC.}~\label{tab:chi-pcc}
	\small
	\begin{tabular}{r >{\centering}m{1.5cm} >{\centering}m{1.5cm} >{\centering}m{1.5cm}}
	    \toprule \midrule
	    $s$=3         & \textsf{dist}         & \textsf{dist-restr}     & \textsf{data-global} \tabularnewline \cmidrule{2-4}
        \textsf{dist-restr}  & 2.29 \quad \quad & - & - \tabularnewline
	    \textsf{data-global} & 95.06$^{**}$     & 95.67$^{**}$ & - \tabularnewline
	    \textsf{data-local}  & 22.32$^{**}$     &  9.22$^{*}$  & 28.29$^{**}$ \tabularnewline
	    \bottomrule
	\end{tabular}
	\\
	\vspace{4mm}
    \begin{tabular}{r >{\centering}m{1.5cm} >{\centering}m{1.5cm} >{\centering}m{1.5cm}}
	    \toprule \midrule
	    $s$=5         & \textsf{dist}         & \textsf{dist-restr}     & \textsf{data-global} \tabularnewline \cmidrule{2-4}
        \textsf{dist-restr}  &  35.59$^{**}$ & - & - \tabularnewline
	    \textsf{data-global} & 327.66$^{**}$ & 165.82$^{**}$ & - \tabularnewline
	    \textsf{data-local}  & 260.92$^{**}$ & 110.79$^{**}$ &  10.69$^{*}$ \tabularnewline
	    \bottomrule
	\end{tabular}
	\\
	\vspace{2mm}
	\raggedleft{\scriptsize
    $^{*}$ $p<0.005$, $^{**}$ $p<0.001$ \qquad
    }
    
    \vspace{6mm}

	\centering
	\small
	\caption{$\chi^2$-values of the McNemar tests for PLV.}~\label{tab:chi-plv}
	\small
	\begin{tabular}{r >{\centering}m{1.5cm} >{\centering}m{1.5cm} >{\centering}m{1.5cm}}
	    \toprule \midrule
	    $s$=3         & \textsf{dist}         & \textsf{dist-restr}    & \textsf{data-global} \tabularnewline \cmidrule{2-4}
        \textsf{dist-restr}  &  4.62$^{*}$ & - & - \tabularnewline
	    \textsf{data-global} & 62.64$^{**}$     & 35.84$^{**}$ & - \tabularnewline
	    \textsf{data-local}  & 17.34$^{**}$     &  5.19$^{*}$  & 12.01$^{**}$ \tabularnewline
	    \bottomrule
	\end{tabular}
	\\
	\vspace{4mm}
    \begin{tabular}{r >{\centering}m{1.5cm} >{\centering}m{1.5cm} >{\centering}m{1.5cm}}
	    \toprule \midrule
	    $s$=5         & \textsf{dist}          & \textsf{dist-restr}     & \textsf{data-global} \tabularnewline \cmidrule{2-4}
        \textsf{dist-restr}  & 133.90$^{**}$ & - & - \tabularnewline
	    \textsf{data-global} & 293.59$^{**}$ & 789.31$^{**}$ & - \tabularnewline
	    \textsf{data-local}  &  69.55$^{**}$ & 346.80$^{**}$ &  61.28$^{**}$ \tabularnewline
	    \bottomrule
	\end{tabular}
	\\
	\vspace{2mm}
	\raggedleft{\scriptsize
    $^{*}$ $p<0.05$, $^{**}$ $p<0.001$ \qquad
    }
\end{table}

The differences between the classification accuracies are statistically evaluated using the McNemar test, which enables pair-wise comparison of the sensitivity and specificity of the classification results obtained using different ordering methods. 
The results ($\chi^2$-values with 1 degree of freedom) are summarized in Tables \ref{tab:chi-pcc} and \ref{tab:chi-plv}. 
It can be observed that the classification performance is significantly different depending on the ordering method, while the only exception is between \textsf{dist} and \textsf{dist-restr} for PCC with a kernel size of 3. 
These results prove that the improvements using the data-driven ordering methods are statistically significant. 
The superiority of \textsf{data-global} over \textsf{data-local} is also revealed as significant in most cases.

\subsection{Discussion}
\label{sec:ord-disc}

\begin{figure*}
    \centering
    \begin{subfigure}{0.9\textwidth}
        \hspace{-6mm}
        \includegraphics[width=\textwidth]{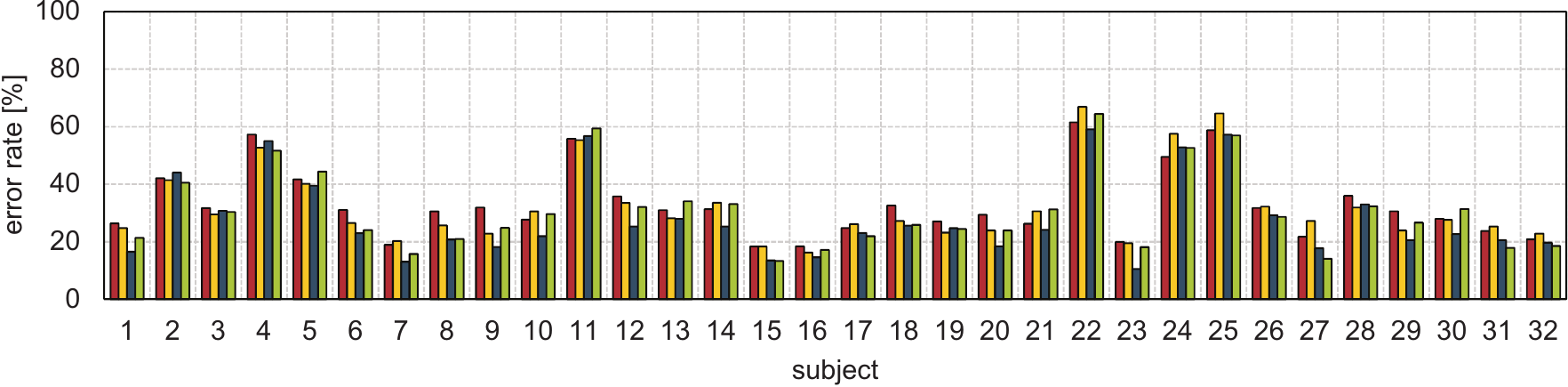}
        \caption{PCC, $s$=3}
        \label{fig:subject-pcc-3}
    \end{subfigure}
    
    \vspace{4mm}
    
    \begin{subfigure}{0.9\textwidth}
        \hspace{-6mm}
        \includegraphics[width=\textwidth]{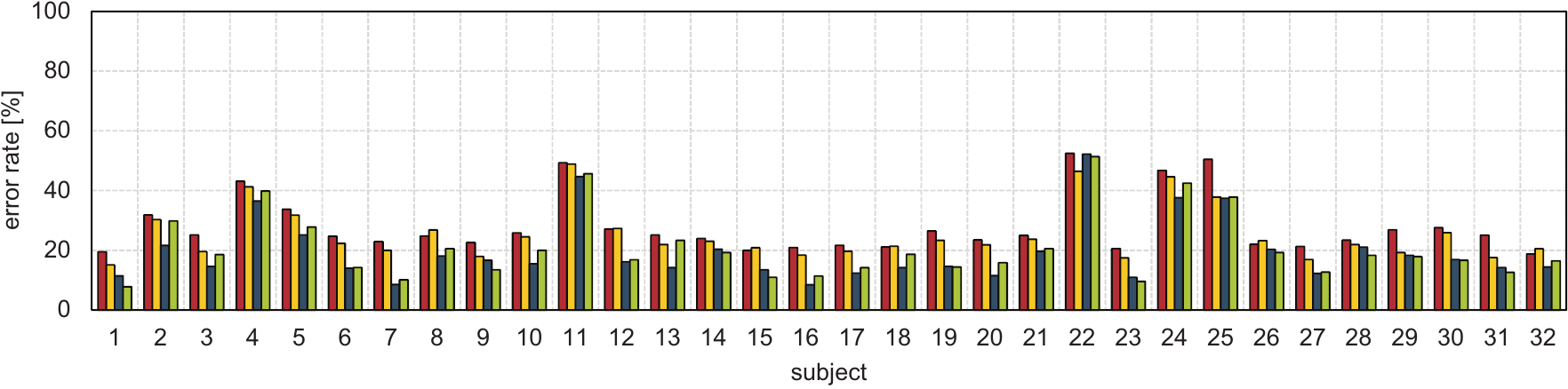}
        \caption{PCC, $s$=5}
        \label{fig:subject-pcc-5}
    \end{subfigure}
    
    \vspace{4mm}

    \centering
    \begin{subfigure}{0.9\textwidth}
        \hspace{-6mm}
        \includegraphics[width=\textwidth]{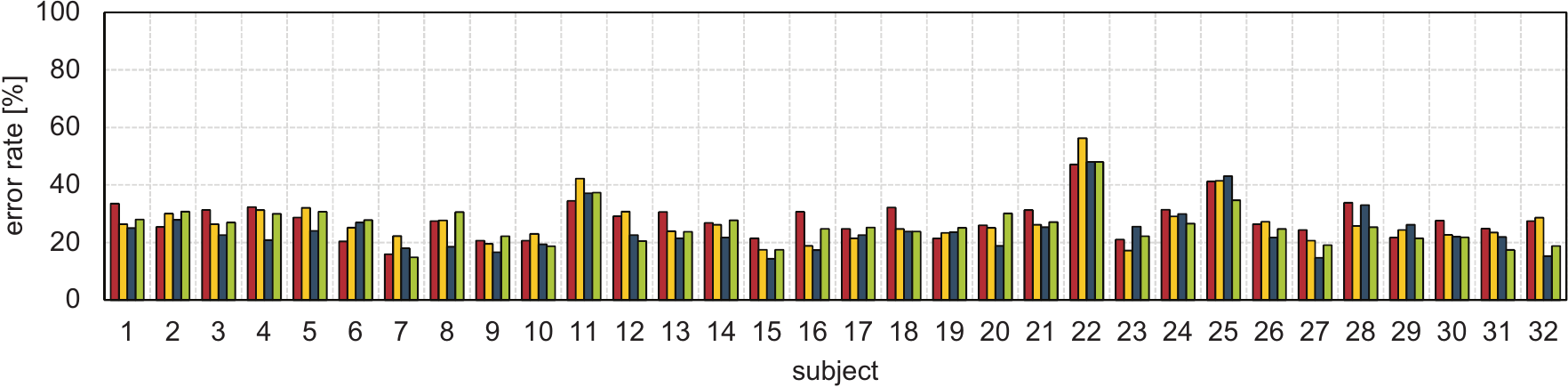}
        \caption{PLV, $s$=3}
        \label{fig:subject-plv-3}
    \end{subfigure}
    
    \vspace{4mm}
        
    \begin{subfigure}{0.9\textwidth}
        \hspace{-6mm}
        \includegraphics[width=\textwidth]{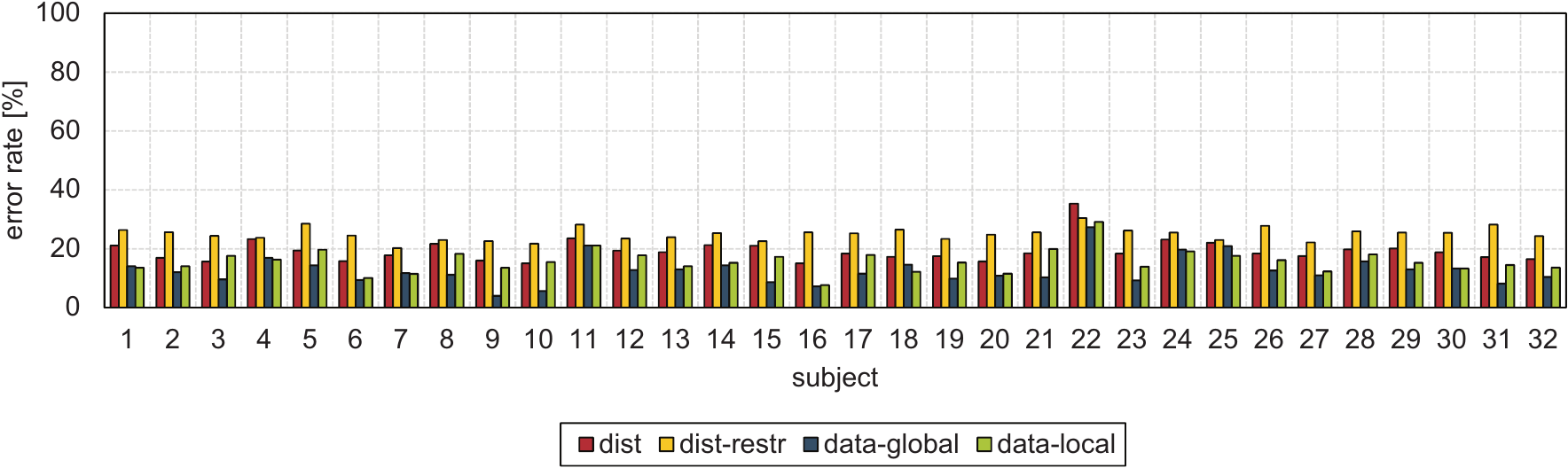}
        
        \vspace{2mm}
        
        \caption{PLV, $s$=5}
        \label{fig:subject-plv-5}
    \end{subfigure}
    
    \caption{Error rates of the emotional video classification depending on the subject.}
    \label{fig:subject}
\end{figure*}

The classification results are further examined in terms of the influence of subjects and videos.  
Figure \ref{fig:subject} shows the classification error rates depending on the subjects.
It is generally known that brain activity significantly varies among individuals, which even enables the identification of individuals based on EEG signals \citep{Delpozo-Banos15}. 
This individual difference appears as large variances in error rates across the subjects, while the difference across the ordering methods is relatively small when the classification performance is relatively poor (Figures \ref{fig:subject-pcc-3} (PCC and $s$=3) and \ref{fig:subject-pcc-5} (PCC and $s$=5)). 
For instance, the standard deviations of the error rates are 13.07\% across the subjects and 2.75\% across the ordering methods in Figure \ref{fig:subject-pcc-3}. 
However, the influence of individuality is reduced as the classification performance improves, which indicates that the improved classification accuracies are obtained by learning the representations of the EEG signals that are robust to the individual difference. 
For example, in Figure \ref{fig:subject-plv-5} (PLV and $s$=5), the standard deviation across the subjects drops to 3.61\%, while that across the ordering methods slightly increases to 4.87\%.

\begin{figure*}
    \centering
    \begin{subfigure}{0.9\textwidth}
        \hspace{-6mm}
        \includegraphics[width=\textwidth]{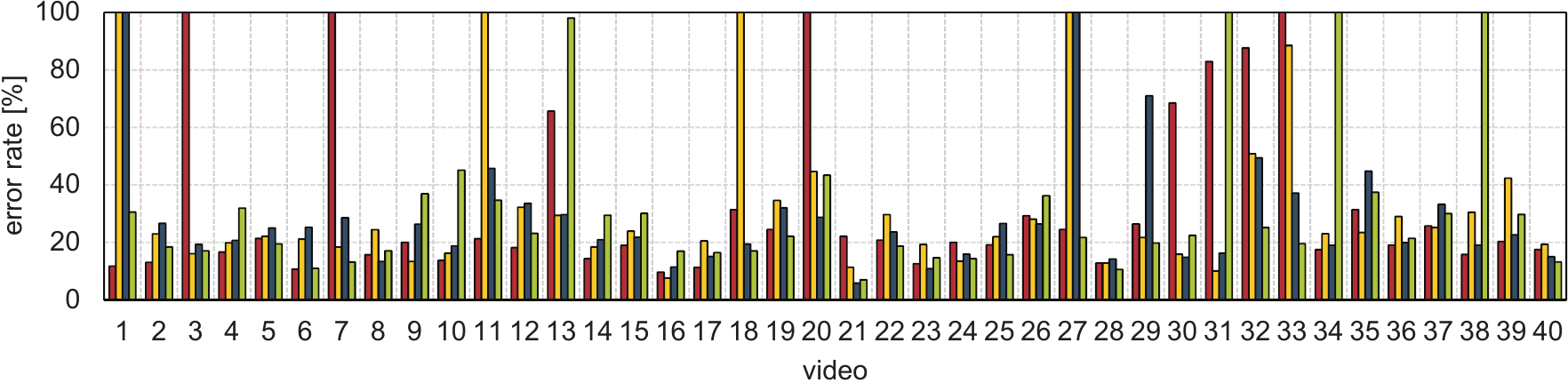}
        \caption{PCC, $s$=3}
        \label{fig:video-pcc-3}
    \end{subfigure}
    
    \vspace{4mm}
    
    \begin{subfigure}{0.9\textwidth}
        \hspace{-6mm}
        \includegraphics[width=\textwidth]{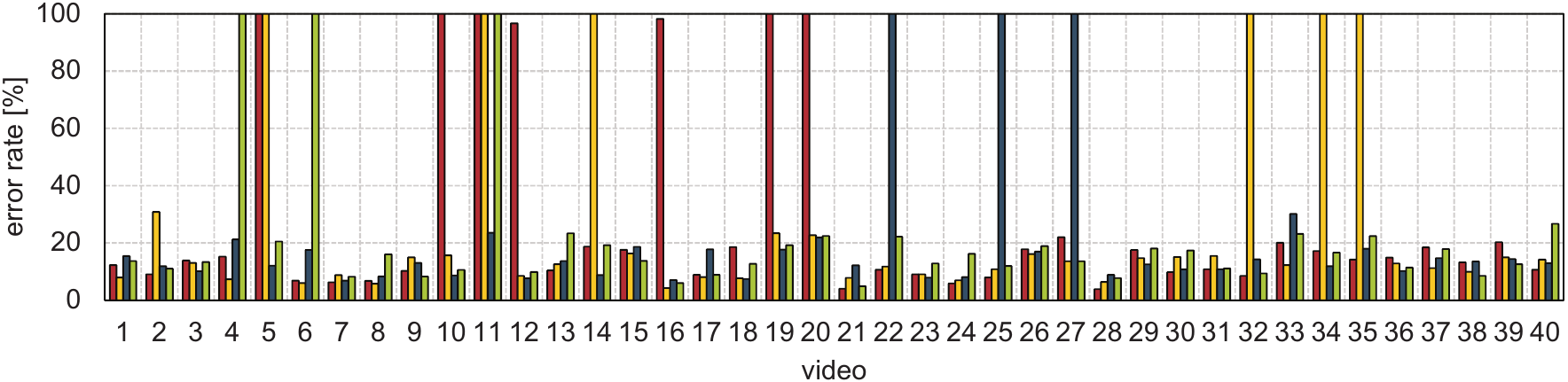}
        \caption{PCC, $s$=5}
        \label{fig:video-pcc-5}
    \end{subfigure}
    
    \vspace{4mm}

    \centering
    \begin{subfigure}{0.9\textwidth}
        \hspace{-6mm}
        \includegraphics[width=\textwidth]{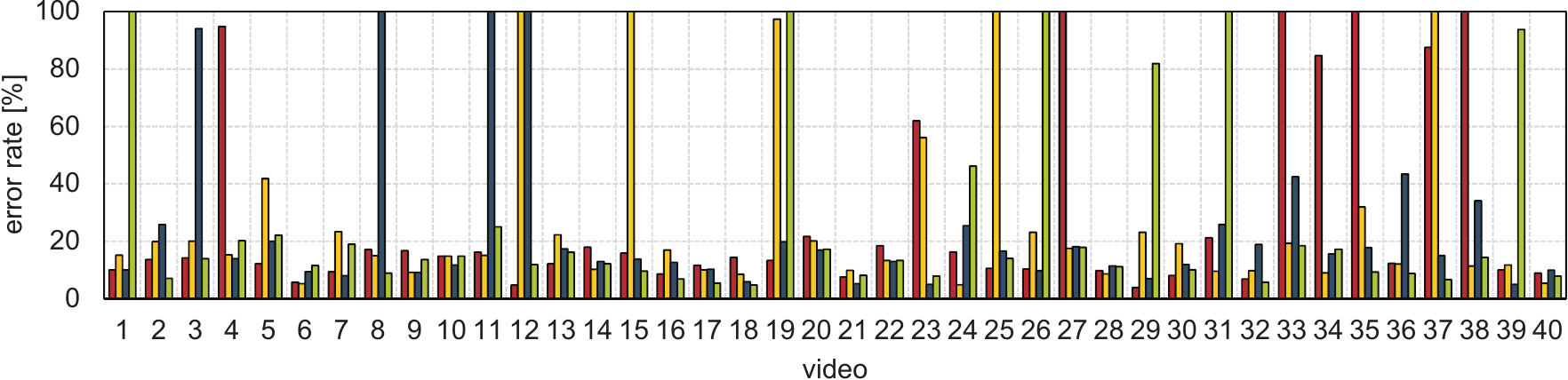}
        \caption{PLV, $s$=3}
        \label{fig:video-plv-3}
    \end{subfigure}
    
    \vspace{4mm}
    
    \begin{subfigure}{0.9\textwidth}
        \hspace{-6mm}
        \includegraphics[width=\textwidth]{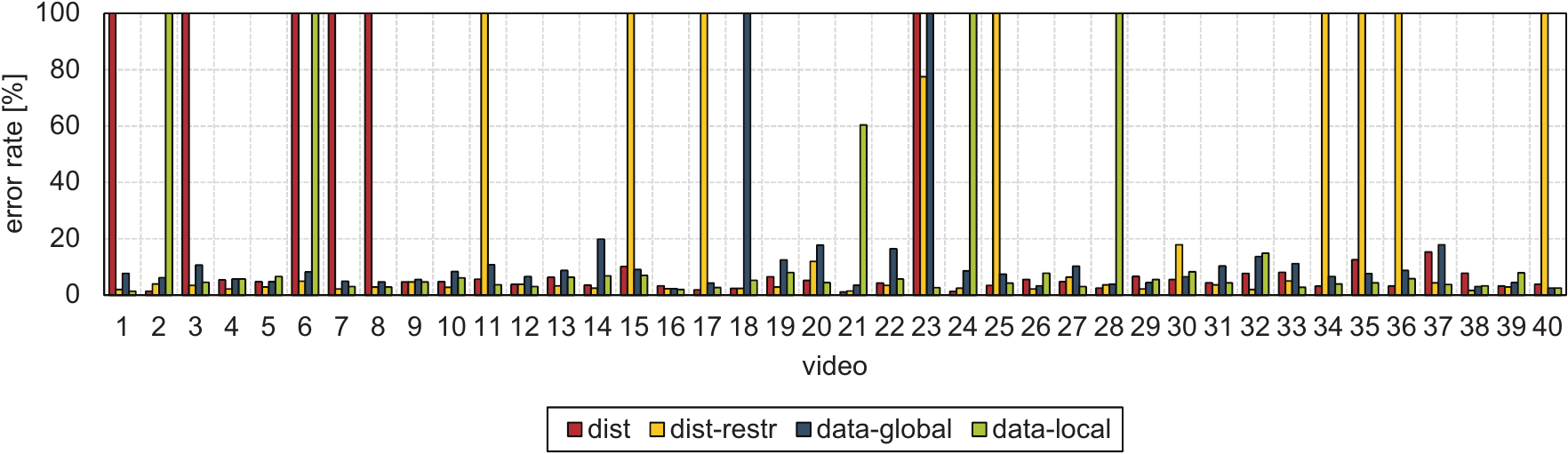}
        
            \vspace{2mm}
            
        \caption{PLV, $s$=5}
        \label{fig:video-plv-5}
    \end{subfigure}
    
    \caption{Error rates of the emotional video classification depending on the video.}
    \label{fig:video}
\end{figure*}

We also analyze the error rates depending on the videos, the results of which are summarized in Figure \ref{fig:video}. 
It is noticeable that the classifiers completely fail to recognize some videos despite well-balanced training and test datasets (each video occupies 2.5$\pm$0.017\% and 2.5$\pm$0.10\% of the entire training and test data, respectively). 
We observed that some of these cases are due to the emotional characteristics of the videos, i.e., videos having similar emotional content are easily misclassified. However, this explains only some cases, and thus we conduct further analysis, as reported in the following section.
% Moreover, the enhancement of classification performance is not achieved by reducing such failure cases but by improving the classification accuracy of the other videos.
% The standard deviations across the videos of Figures \ref{fig:video-pcc-3}-\ref{fig:video-plv-5} are 24.95\%, 27.85\%, 29.50\%, and 30.80\%, respectively. 

\subsection{Concentration of emotion-related connectivity}
\label{sec:concent}

It is noteworthy that the videos for which the classification is unsuccessful are not consistent, even between the data-driven ordering methods (Figure \ref{fig:video}). 
This indicates that such failures are cased by the characteristics of the videos and classification systems interactively resulting in significantly different classification accuracies across the videos rather than by defects in the videos. 

Therefore, we evaluate the classification performance by concurrently considering the characteristics of the videos and classification systems. 
In other words, we consider that the properties of classification systems, i.e., the type of connectivity measure, the kernel size, and the ordering method, affect the recognition performance of videos with specific emotional characteristics. 
Specifically, we hypothesize that it is advantageous for classification when the electrode pairs related to the valence of target videos are placed close together in the connectivity matrix.
The valence is one of the major bases of emotional states, and the concentration of functionally similar connectivity makes it easier to capture distinguishing features for classification by convolution operations.

\begin{table}[t]
	\centering
	\small
	\caption{Error rates (\%) of the emotional video classification depending on the valence.}~\label{tab:valence}
	\begin{tabular}{r c c c c c}
		\toprule \midrule
		\multirow{2}{*}{Ordering}               & \multirow{2}{*}{Valence} & \multicolumn{2}{c}{PCC} & \multicolumn{2}{c}{PLV} \\ 
		                                        & & $s$=3   & $s$=5   & $s$=3   & $s$=5 \\ \midrule
		\multirow{2}{*}{\textsf{data-global}}   & low  & 29.18 & 26.12 & 17.57 & 12.25 \\
		                                        & high & 27.81 & 13.48 & 30.50 & 13.02 \\ 
		\multirow{2}{*}{\textsf{data-local}}    & low  & 33.36 & 15.31 & 29.77 & 17.56 \\
		                                        & high & 28.47 & 26.62 & 21.91 & 13.96 \\ 
		\bottomrule
	\end{tabular}
\end{table}

\subsubsection{Valence vs. classification performance}
\label{sec:concent-valence}

The error rates of the data-driven ordering methods are examined depending on the valence of the target videos to show that the valence can explain the varying classification accuracies with the videos (Table \ref{tab:valence}). 
It can be seen that the error rates significantly differ for the high- and low-valence videos despite the balanced distribution of valence in the test dataset (49.4\% and 50.6\% for high- and low-valence videos, respectively). 
Better classification performance is observed for the high-valence videos with the PCC connectivity matrices in most cases except when using the \textsf{data-local} method with a kernel size of 5. 
The results obtained using PLV exhibit a different tendency from those using PCC: 
\textsf{data-global} produces lower error rates for the low-valence videos whereas \textsf{data-local} shows better classification performance for the high-valence videos. 

\begin{table*}[t]
    \centering
    \small
	\caption{Electrode pairs corresponding to low- and high-valence-related connectivity \citep{Lee14,Martini12}.}~\label{tab:conn}
    \begin{tabular}{r >{\raggedright}p{.8\columnwidth} >{\raggedright}p{.8\columnwidth}}
	    \toprule \midrule
	     & Low-valence & High-valence \tabularnewline \midrule
	    PCC & 
	    Fp2-F7, F7-O1, Fz-T8, T8-P3, P8-O2, O1-O2 & 
	    Fp1-F8, Fp1-P7, Fp2-C3, F3-F4, F3-P7, F4-C3, F7-F8, F7-P8, Fz-T8, C3-T8, T7-T8 \tabularnewline
	     & & \tabularnewline
	    PLV &
	    Fp1-Fp2, Fp1-FC1, Fp1-FC2, Fp1-F4, Fp1-Fz, Fp1-Cz, Fp1-P8, Fp2-Fz, Fp2-P3, F3-Fz, F4-T8, F8-Fz, F8-P7, Fz-P4, T7-P4, T7-P7, T7-Pz, T8-P7, T8-Pz, C3-P3, C3-P4, P7-O2 &
	    Fp2-T7, Fp2-Pz, F3-Fz, Fz-C4, Fz-P4, C3-C4, C4-P3, Cz-Pz, T7-Pz, P7-P8, P7-O2 \tabularnewline
	    \bottomrule
	\end{tabular}
\end{table*}

\subsubsection{Concentrativeness}
\label{sec:concent-measure}

Previous studies have analyzed which electrode pairs form connectivity related to valence. 
\cite{Lee14} identified valence-related PCC and phase synchronization index (PSI); the latter is a measure of phase synchronization and thus can be equated to PLV. 
Meanwhile, valence-related PLV was revealed by \cite{Martini12}. 
Low- and high-valence-related connectivity incidences based on these studies are summarized in Table \ref{tab:conn}, in which the connectivity incidences that showed a significant difference for either negative-neutral or positive-neutral valence are considered as low- or high-valence-related connectivity, respectively. 

We propose a measure to quantify the degree of concentration of valence-related connectivity in a connectivity matrix, called \textit{concentrativeness}. 
It is defined as the ratio of the number of connectivity incidences of interest (low- or high-valence-related connectivity) to the number of entire connectivity incidences included in the receptive field (a patch of the connectivity matrix captured by the sliding window of a convolutional filter), i.e.,

\begin{equation}
    C = \frac{1}{N-M}\sum_{n=1}^{N}r_n.
\end{equation}
Here, $r_n$ indicates the ratio of the connectivity incidences of interest in the $n$-th sliding window of a convolutional filter (e.g., when four elements in the $n$-th sliding window of a $3\times3$ convolutional filer are among the low-valence-related connectivity incidences listed in Table \ref{tab:conn}, the value of $r_n$ for low-valence becomes $\frac{4}{3\times3}$),
$N$ is the total number of sliding windows (e.g., $N=32\times32=1024$ for the first convolutional layer of the CNN architecture used in our experiment), and  
$M$ indicates the number of sliding windows that do not include a connectivity incidence of interest. 
In other words, the sliding windows with a corresponding $r_n$ value of zero are excluded from the calculation of the average ratio of the connectivity incidences of interest.
Therefore, $\frac{1}{N-M}$ acts as a normalization factor to remove the influence of different numbers of low- and high-valence-related connectivity incidences.

\begin{figure}[t]
    \centering
    \begin{subfigure}{0.23\textwidth}
        \hspace{-6mm}
        \includegraphics[width=\textwidth]{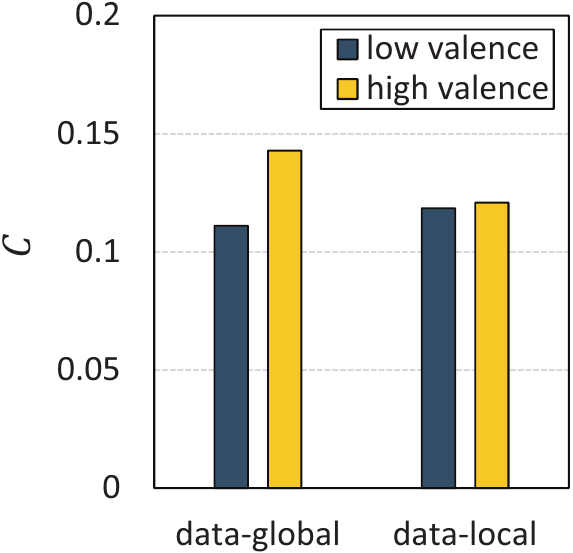}
        \caption{PCC, $s$=3}
        \label{fig:conn-pcc-3}
    \end{subfigure}
    \begin{subfigure}{0.23\textwidth}
        \hspace{-6mm}
        \includegraphics[width=\textwidth]{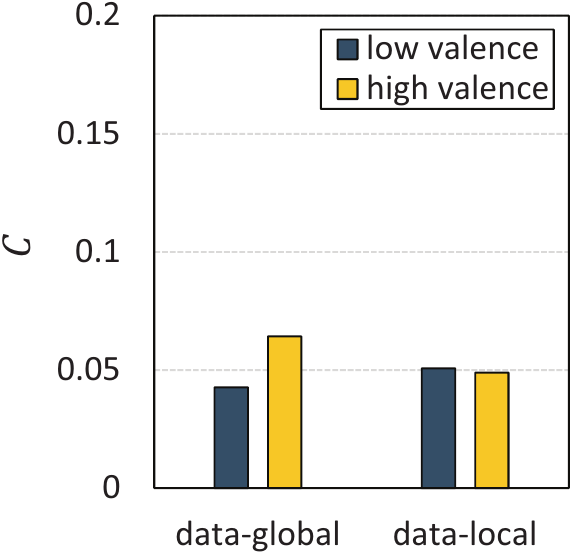}
        \caption{PCC, $s$=5}
        \label{fig:conn-pcc-5}
    \end{subfigure}
    
    \vspace{3mm}
    
    \begin{subfigure}{0.23\textwidth}
        \hspace{-6mm}
        \includegraphics[width=\textwidth]{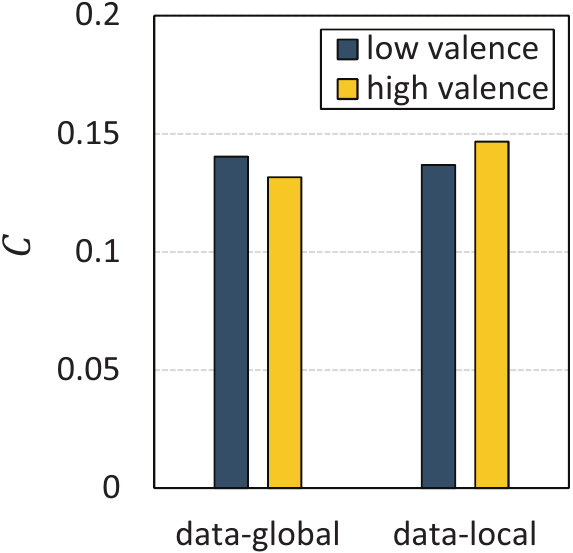}
        \caption{PLV, $s$=3}
        \label{fig:conn-plv-3}
    \end{subfigure}
    \begin{subfigure}{0.23\textwidth}
        \hspace{-6mm}
        \includegraphics[width=\textwidth]{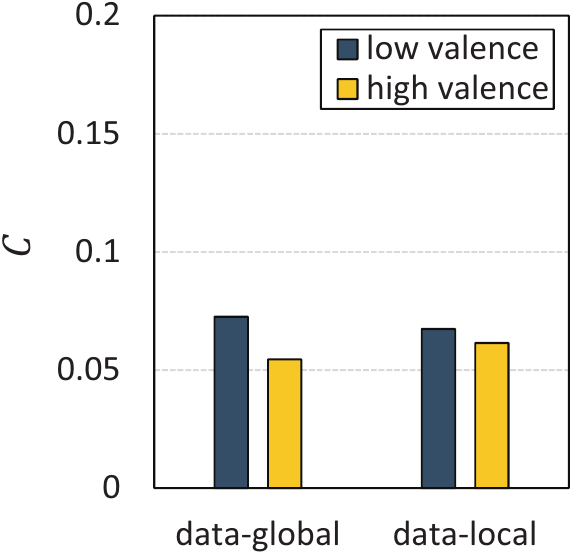}
        \caption{PLV, $s$=5}
        \label{fig:conn-plv-5}
    \end{subfigure}
    
    \caption{Valence-related concentrativenesses of the \textsf{data-global} and \textsf{data-local} ordering methods.}
    \label{fig:conn}
\end{figure}

Figure \ref{fig:conn} shows the concentrativenesses of the low- and high-valence-related connectivity incidences at the first convolutional layer. 
The concentrativeness is negatively correlated with the error rate presented in Table \ref{tab:valence}, which indicates that our assumption is verified. 
The Spearman's rank-order correlation coefficient between the error rates and concentrativenesses is $-0.667$, meaning that the classification performance is enhanced by utilizing connectivity matrices where the connectivity incidences related to the valences of the target videos are more concentrated and thus the activation values via convolution operations become more distinct. 
And, this effect results in the varying classification accuracies depending on the video in Figure \ref{fig:video}.

%%%%%%%%%%%%%%%%%%%%%%%%%%%%
\section{Conclusions}
\label{sec:conclusion}

A new approach to utilizing brain connectivity via EEG signals using CNNs was proposed. 
We demonstrated the significance of the proposed method for the emotional video classification task and compared three connectivity measures that reflect different aspects of brain connectivity. 
Moreover, the data-driven methods, introduced for the optimal arrangement of the connectivity matrix, significantly improved the classification performance compared with the connectivity matrix arrangement based on the locations of the EEG electrodes. 
We conducted further analysis to explain the classification results and clarified the influence of the arrangement of the connectivity matrix on the classification performance. 
The performance significantly differed depending on the target class (i.e., video), which was explained by the valence of emotion induced by the videos. 
We found that the classification performance for low- or high-valence videos is correlated with the concentrativeness of the related connectivity incidences in the connectivity matrix. 
That is to say, the distinguishing features for emotional video classification are caught effectively when the related connectivity incidences are closely located. 

Although we dealt with the emotional video classification task in this work, we believe that the proposed approach could also be successfully applied to other emotional state classification and prediction tasks.
Our results in Section \ref{sec:connmat-result} and our preliminary results in \cite{Moon18} confirm the effectiveness of CNNs using connectivity features for emotional video classification and video-induced valence classification, respectively.
Furthermore, our analysis in Section \ref{sec:concent} demonstrates that the data-driven ordering methods on top of CNNs with connectivity features are successful largely due to the effective processing of connectivity patterns related to emotion (i.e., valence).

In the future, the effectiveness of the connectivity matrix and data-driven ordering methods will be validated for different tasks. 
It will be also interesting to apply our method to other brain imaging modalities such as fMRI, MEG, and fNIRS. 


\section*{Acknowledgment}
The research of Seong-Eun Moon and Jong-Seok Lee was supported by the MSIT (Ministry of Science and ICT), Korea, under the ``ICT Consilience Creative Program'' (IITP-2019-2017-0-01015) supervised by the IITP (Institute for Information \& communications Technology Promotion), and the research of Jane-Ling Wang was supported by the US National Science Foundation DMS-19-14917.

\bibliographystyle{apa} 
\bibliography{refs.bib}


%\begin{thebibliography}{00}
%
%%% \bibitem[Author(year)]{label}
%%% Text of bibliographic item
%\end{thebibliography}
\end{document}